\documentclass[conference]{IEEEtran}
\usepackage{amsmath,graphicx}
\usepackage{flushend}
\usepackage{multirow}
\usepackage[bookmarks=false]{hyperref}
\usepackage[all]{hypcap}

\ifCLASSINFOpdf
\else
\fi
\begin{document}
%
\title{A 2.5D Cascaded Convolutional Neural Network with Temporal Information for Automatic Mitotic Cell Detection in 4D Microscopic Images}



%
\author{\IEEEauthorblockN{Titinunt Kitrungrotsakul\IEEEauthorrefmark{1},
Xian-Hau Han\IEEEauthorrefmark{2},
Yutaro Iwamoto\IEEEauthorrefmark{1},
Satoko Takemoto\IEEEauthorrefmark{3},
Hideo Yokota\IEEEauthorrefmark{3},\\
Sari Ipponjima\IEEEauthorrefmark{4},
Tomomi Nemoto\IEEEauthorrefmark{4},
Xiong Wei\IEEEauthorrefmark{5} and
Yen-Wei Chen\IEEEauthorrefmark{1}}
\IEEEauthorblockA{\IEEEauthorrefmark{1}Graduate School of Information Sci. and Eng., Ritsumeikan Univ., Shiga 525-8577, Japan}
\IEEEauthorblockA{\IEEEauthorrefmark{2}Faculty of Science, Yamaguchi University, Yamaguchi, Japan}
\IEEEauthorblockA{\IEEEauthorrefmark{3}Center for Advanced Photonics, RIKEN, Saitama, Japan}
\IEEEauthorblockA{\IEEEauthorrefmark{4}Research Institute for Electronic Science, Hokkaido University, Hokkaido, Japan}
\IEEEauthorblockA{\IEEEauthorrefmark{5}Institute for Infocomm Research, Singapore}}


\maketitle

\begin{abstract}
In recent years, intravital skin imaging has been increasingly used in mammalian skin research to investigate cell behaviors. A fundamental step of the investigation is mitotic cell (cell division) detection. Because of the complex backgrounds (normal cells), the majority of the existing methods cause several false positives. In this paper, we proposed a 2.5D cascaded end-to-end convolutional neural network (CasDetNet) with temporal information to accurately detect automatic mitotic cell in 4D microscopic images with few training data. The CasDetNet consists of two 2.5D networks. The first one is used for detecting candidate cells with only volume information and the second one, containing temporal information, for reducing false positive and adding mitotic cells that were missed in the first step. The experimental results show that our CasDetNet can achieve higher precision and recall compared to other state-of-the-art methods.

\end{abstract}

\renewcommand\IEEEkeywordsname{Keywords}

\vspace{0.3cm}

\section{Introduction}

\begin{figure}[t!]

\begin{minipage}[b]{1.0\linewidth}
  \centering
  \centerline{\includegraphics[width=3cm]{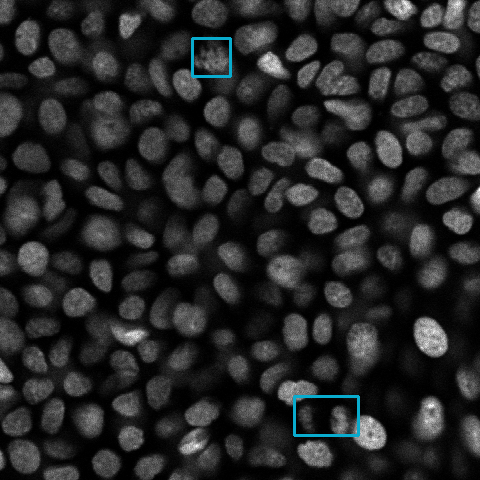}}
\end{minipage}
\caption{One typical slice image of an observed 3D live cell dynamic image(4D image) and mitotic cells (cell division) are indicated by blue bounding boxes.}
\label{fig:ori}
\end{figure}

Division of the cell in adult mammalian epidermis is important for maintaining the epidermal structure as these cells are important for replenishing eliminated keratinocytes \cite{Sari2016}. Cancer, atopic dermatitis, ichthyosis vulgaris, and skin diseases disrupt the balance between the proliferation and elimination of keratinocytes and create abnormal skin structures \cite{Hsu14,Jones08,Watt14}. Though detecting the mitotic cell (cell division) is essential in investigating cell behaviors, the majority of the methods and experiments were performed with 2D dynamic images that may overlook the important information can result in wrong detection. 3D live cell dynamic images (4D images) can be obtained by using a two-photon microscopy \cite{Sari2016}. A typical slice image of an observed 3D dynamic image is shown in Fig.\ref{fig:ori}, with blue bounding boxes indicating the mitotic cells (cell division). Automatic detection of mitotic cells from such 3D dynamic images (4D images) is a challenging task. Recently, deep learning architecture has demonstrated the powerful ability of computer vision tasks by automatically learning hierarchies of relevant features directly from the input data. The deep convolutional neural network has been successfully applied for image classification and object detection, especially for ImageNet classification competition, which has been the most successful network for image classification since 2012 \cite{Krizhevsky12}. Moreover, Fast Region-based Convolutional Networks (Fast R-CNN) for object detection and Single Shot MultiBox Detector (SSD) are powerful methods, both of which have outperformed several other methods, that use CNN as base network to perform object detection \cite{FRCNN, SSD}. However, these methods are designed for 2D natural image detection. In the field of mitotic cell detection, varies methods have been proposed, most of which are based on image binarization \cite{Sauvola2000} or segmentation of cells \cite{Meijering2012}. Though those methods is that they do not require training dataset to train the model, they require proper alignment between each slice or time sequence to obtain good results, which is time-consuming. Anat et al. \cite{Anat15} used a deep learning method called pixel-wised method to improve detection accuracy and accelerate the computation time. This method is based on 2D patch classification using a simple CNN network and takes considerable computation time. Though we can apply Fast R-CNN and SSD, which are widely used for object detection in natural images, they will cause several false positives because the object (mitotic cell) is similar to the background image (normal cells), as shown in Fig.\ref{fig:ori}. In this paper, we proposed a 2.5D cascaded end-to-end convolutional neural network (CasDetNet) with temporal information for accurate automatic detection of 4D (x, y, z, t) mitotic cell division events in epidermal basal cells with few training data. The CasDetNet consists of two 2.5D networks. The first is used for detecting candidate cells with only volume information and the second one, with temporal information, is used for reducing false positives (normal cells) and adding mitotic cells that are missing in the first step. We also intend to use a 2.5D CNN as a base network. Compared to conventional 2D CNN, our 2.5D CNN (2D image with neighbor slices) can include more information for detection (the first step) and reduction of false positives (the second step). Though the 3D CNN can include more information than 2D and 2.5D CNN, it can use limited number of training samples (3D images) and thus cause overfitting. Results show that CasDetNet can deliver higher precision and recall comparing to other advanced methods.

The paper is organized as follows. Section \ref{sec:method} introduces the proposed CasDetNet for mitotic cell detection method is introduced in section \ref{sec:method}. Section \ref{sec:results} describes the experimental results. Finally, Section \ref{sec:Conclusion} presents the conclusion.

\begin{figure}[t!]

\begin{minipage}[b]{1.0\linewidth}
  \centering
  \centerline{\includegraphics[width=7.0cm]{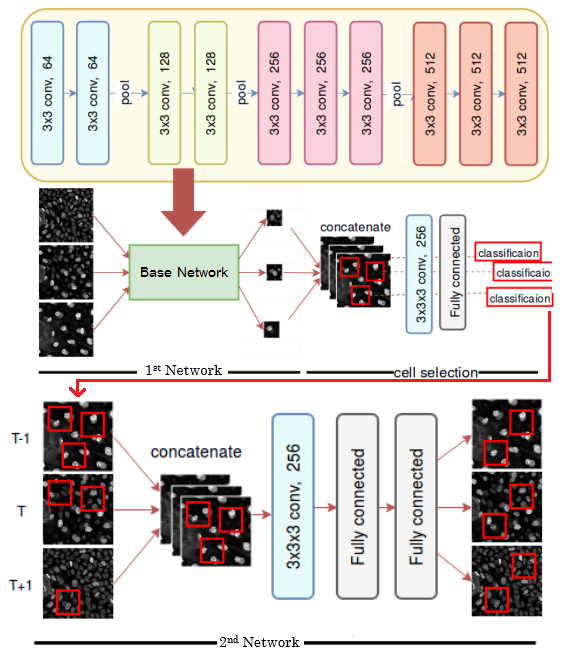}}
\end{minipage}
\caption{Overview of our proposed CasDetNet.}
\label{fig:based}
\end{figure}

\section{The Proposed Network}
\label{sec:method}

The proposed CasDetNet for detection of mitotic cells is shown in Fig.\ref{fig:based}. It comprises two 2.5D networks. The first network is used to detect candidate cells using only volume information and the second, which contains temporal information, is used to reduce false positives and to add mitotic cells that were missing in the first step. The second network is cascaded to the first network and the two networks are then trained simultaneously (end-to-end training). The details regarding the first and second networks will be described in subsections \ref{ssec:selector} and \ref{ssec:refine}, respectively.

\subsection{The first network for detection of candidate cells using volume information}
\label{ssec:selector}

%
%

The first network for detecting candidate cells is motivated by Fast R-CNN to determine the local features for establishing the region of interest (ROI). The goal is to cause the network's hidden layers to detect candidate of mitotic cells. The original Fast R-CNN requires 2D image as input and produces a set of ROI as detection results. The number of training set and network architectures determine the quality of detection result. Further, the conventional Fast R-CNN's drawback is that it loses 3D spatial information, which is important for accurate mitotic cell detection. Though we can extend the conventional Fast R-CNN to a 3D version for 3D volume images, the number of training samples will be considerably limited and result in over-fitting. Thus, we propose a 2.5D Fast R-CNN for our first detection network. As shown in Fig.\ref{fig:ori}, three slice images ${\{s_{-1},s,s_{+1}\}}$ are used as input to detect the candidate cells in the target slice image ${\{s\}}$, which is called 2.5D network. The outputs (ROIs) are indicated as ${\{o_1,o_2,o_3,...,and so on\}}$. The advantage of our 2.5D network is that we can use neighbor slice information (2.5D information) to distinguish between the mitotic cell and normal cells, which is important for detecting mitotic cells divided along z-axis.

\begin{figure*}[t!]
\begin{minipage}[b]{1.0\linewidth}
  \centering
  \centerline{\includegraphics[width=15cm]{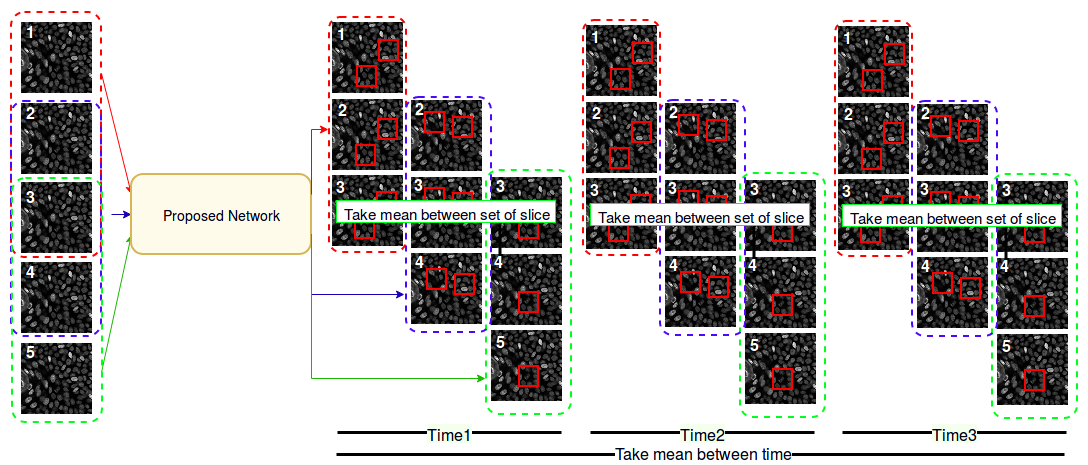}}
\end{minipage}
\caption{Taking mean result to eliminate normal cell in detected result from both volume and time information detection network.}
\label{fig:voting}
\end{figure*}

Figure \ref{fig:based} (upper part) illustrates our first 2.5D network. For our base network, we use the VGG network architecture. To enhance the accuracy of the network, we use transfer learning from ImageNet data to VGG. Each slice is processed individually and the processed slices are concatenated to form a 3D volume. We replace 3D convolutional layer with 3 ${\times}$ 3${\times}$ 3 kernel size, followed by a ReLU non-linearity layer instead of 2D convolutional layer of original Fast R-CNN to obtain network results ${o_{i}}$. In each output ${o_{i}}$, the network will use nearby output ${o_{-1}}$ and ${o_{+1}}$ to generate the concatenated output. It will also be used in cell selection process to generate volume selected output ${O_i^t}$. Thus, the network will generate a set of output ${O_i^t}$ consisting of 3 outputs ${\{o_{i1},o_{i2}, o_{i3}\}^{t}}$ for each image slice ${s_i}$. Further, ${t}$ indicates the time sequence in 4D data. For each set of output ${O_i^t}$, we calculate the mean to obtain first volume output ${V_i^t}$, as shown in Fig.\ref{fig:voting}.


\subsection{The second Network for reduction of false positive}

\label{ssec:refine}

We propose using the second network to reduce the false positives generated by the first network. Results from the first network ${V_i^t}$ contain both correct and incorrect detection results. In this section, we used the second network to refine the results (reduce the false positives) by using temporal ${\{time_1,time_2,time_3,...,time_n\}}$ information.

There are several methods to manage extra dimensional information (temporal information). Taking mean or thresholding from image sequence is a common method to smoothen the image sequence and removing or adding over-/under-detection. We concatenate the volume output ${V_i^t}$ at time ${t}$, previous output ${V_i^{t-1}}$, and next output ${V_i^{t+1}}$ time sequence together and then apply the second CNN classification (for reducing false positives), as shown in Fig.\ref{fig:based} (lower part). The network will generate the time set of output consisting of three outputs of temporal frames ${\{time_i^{t-1}, time_i^{t}, time_i^{t+1}\}}$. The final result ${F_i^t}$ is obtained by taking the mean of three frames, as shown in Fig.\ref{fig:voting}.

\begin{table}[t!]
\caption{Detection performance of our proposed CasDetNet on 2D slice image.}
\begin{center}
\begin{tabular}{ l | c  c  c  }
  \hline			
  Data & true positive & false positive & ground truth\\ \hline
  1 & 662 & 183 & 711  \\
  2 & 628 & 756 & 745 \\
  3 & 1296 & 216 & 1717 \\
  4 & 1215 & 895 & 1576 \\
  5 & 183 & 229 & 1563 \\
  \hline
\end{tabular}
\end{center}
\label{table:raw}
\end{table}

\section{Experimental results}
\label{sec:results}

 To validate the effectiveness of our proposed method, we perform experiments on 4D (temporal 3D volume sequence) data from JSPE, Technical committee on Industrial Application of Image Processing Appearance inspection algorithm contest 2017 (TC-IAIP AIA2017) \cite{ViEW}. There are five datasets, each containing approximately 80 temporal frames. The data size is approximately 480${\times}$480${\times}$37. Each data contains 1–3 mitotic cells, as listed in Table \ref{table:application} (ground truth). Data augmentation is added in the training phase to increase the number of training set so that overfitting that normally occurs in small datasets can be avoided and the model can be induced to learn to detect the mitotic cells that will generally be under-detected in the 2D network. Cropping, rotation, translation, mirror imaging, noising, and resizing methods are used in our study. The parameters for cropping, rotation, translation, noising, and resizing are randomly selected. We determine the parameter for each augmentation method as follows: 224${\times}$224 cropping size with random location; random rotation angle in the range of 0–180; random percentage of Gaussian noise in the range of 1${\%}$-3${\%}$; random resizing scale in the range of 0.9–1.1. Using data augmentation methods helps to generate varied combination images to train the model.

\begin{figure}[t!]

\begin{minipage}[b]{1.0\linewidth}
  \centering
  \centerline{\includegraphics[width=7cm]{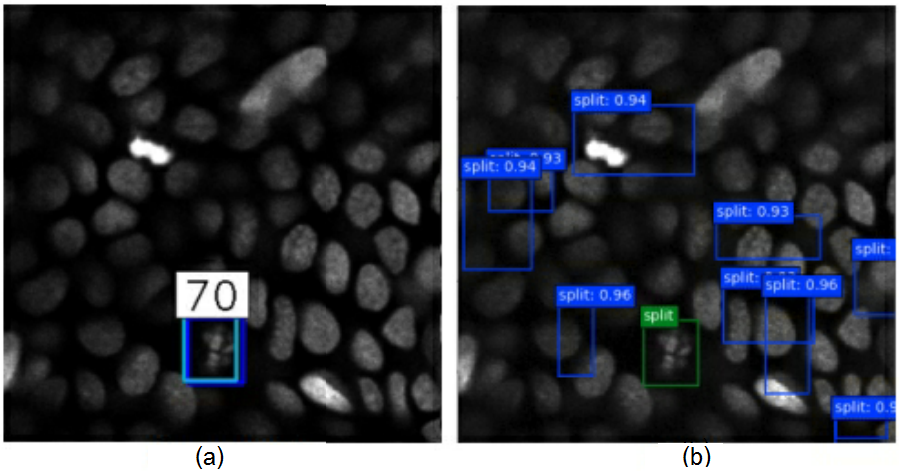}}
\end{minipage}

\caption{Typical detection results on 2D slice image by our proposed CasDetNet (a) and SSD (b).}
\label{fig:visualization}

\end{figure}

\begin{table}[t!]
\caption{Quantative comparison of our proposed CasDetNet with the state-of-the-art methods on 2D slice images.}
\begin{center}
\begin{tabular}{ l | c | c | c }
  \hline			
  Method & precision & recall & time(sec) \\ \hline
  2D FAST R-CNN  \cite{FRCNN} & 0.0870 & 0.9310 & 239.409 \\
  3D FAST R-CNN  & 0.0592 & 0.4143 & 1989.012 \\
  SSD \cite{SSD} & 0.0411 & 0.7221 & 102.551 \\\hline
  Our first network & 0.3591 & 0.7532 & 253.005 \\
  CasDetNet & 0.7228 & 0.70358 & 329.771 \\
  \hline
\end{tabular}
\end{center}
\label{table:comparison}
\end{table}

In our experiments, we use leave-one-out method. Further, for training our model, we use Adam optimization method. As described in the previous section, two networks are cascaded and trained simultaneously (end-to-end). The learning rate for Adam in our network starts with ${0.5 \times 10^{-5}}$ and changes into ${0.5 \times 10^{-6}}$ after finishing the 10k batch, with each batch containing five image slices.

\subsection{Detection results on 2D slice images}
\label{ssec:result2d}

First, we present detection results on 2D slice images. Each slice image is considered as a sample. The total number of mitotic cells (2D slice images) is shown in Table I as ground truth, and precision and recall are used as quantitative measures. For evaluation, we compare the precision and recall of our method with SSD \cite{SSD}, FAST R-CNN \cite{FRCNN}, and 3D convolution FAST R-CNN, which is a modified version of the original FAST R-CNN. All methods are calibrated from ImageNet except 3D FAST R-CNN. The detection results for 2D slice images using CasDetNet are shown in Table \ref{table:raw}. The number of true positive ROI of all data is largely the same as the number of ground truth ROI, except for Data No. 5 that cannot be detected properly as its mitotic cells were difficult to detect because they occur at the edge of the image. The detection results for 2D slice images obtained using our proposed method and SSD are shown in Fig.\ref{fig:visualization}. It is evident that our method can detect mitotic cell correctly. On the other hand, several false positives are detected by SSD (Fig.\ref{fig:visualization}(b)). Compared to the SSD result (Fig.\ref{fig:visualization}(b)), our proposed method (Fig.\ref{fig:visualization}(a)) can significantly reduce false positives. The quantitative comparisons are shown in Table II. Though both 2D FAST R-CNN and SSD present high recall, they also present low precision because of several false positives being detected. Both precision and recall for 3D FAST R-CNN are lower because of overfitting. The 3D FAST R-CNN also has high computation cost. If we only use the first 2.5D network, we can improve the precision compared to 2D FAST R-CNN and SSD because of the 2.5 D network. However, it still contains a large number of false positives. We can also significantly reduce these false positives by using the second network with temporal information. It should be noted that we do not compare our method with Anat's method \cite{Anat15} because it is a pixel-wise method and takes more time than 3D FAST R-CNN in both training and testing.

\subsection{Detection results on 4D data}
\label{ssec:result4d}

Our aim is to detect mitotic cells on 4D data. We combine our detection results on 2D slice image, as described in the previous sub-section, for final results and compare our results with the winner of the TC-IAIP AIA2017 contest \cite{ViEW}. The detection results (TP, FN, FP) regarding 4D data are summarized in Table III. Except Data No. 5, perfect detection is achieved without any FP and FN. For Data No. 5, two mitotic cells are not detected, the reason for which has been described in the previous sub-section. Sugano method \cite{Junichi2017}, the winner in TC-IAIP AIA2017 contest, can also properly detect mitotic cells. However, there are 3 FP for Data No. 2.

\begin{table}[t!]
\caption{Detection results on 4D images.}
\begin{center}
\begin{tabular}{ c | c | c | c  | c | c | c | c }
  \hline			
\multirow{2}{*}{Data} & \multicolumn{3}{c|}{Sugano\cite{Junichi2017}} & %
    \multicolumn{3}{c|}{Our method} & \multirow{2}{*}{ground truth}\\
\cline{2-7}

  & TP & FN & FP & TP & FN & FP & \\ \hline
  1 & 1 & 0 & 0 & 1 & 0 & 0 & 1 \\
  2 & 1 & 0 & 3 & 1 & 0 & 0 & 1 \\
  3 & 2 & 0 & 0 & 2 & 0 & 0 & 2 \\
  4 & 3 & 0 & 0 & 3 & 0 & 0 & 3 \\
  5 & 2 & 1 & 0 & 1 & 2 & 0 & 3 \\

  \hline
\end{tabular}
\end{center}

\label{table:application}

\end{table}

\section{Conclusion}
\label{sec:Conclusion}

We have proposed a 2.5D cascaded convolutional neural network for automatic detection of mitotic cells in 4D image (x, y, z, and time). The proposed network consists of two networks, the first of which is a modified 2.5D Fast R-CNN for detecting candidate cells and the second is used for reducing false positives using temporal information. The results demonstrated that our proposed method is more accurate than the other established methods such as Fast R-CNN, SSD, and the TC-IAIP AIA2017 contest winners' method.


\section*{Acknowledgement}
\label{sec:Acknowledgement}

This work is supported in part by Japan Society for Promotion of Science (JSPS) under Grant No. 16J09596 and KAKEN under the Grant No. 18H04747, 16H01436, 15H05954, 15H05953 and also partially supported by A*STAR Research Attachment Programme.

%
\IEEEpeerreviewmaketitle



%

\bibliographystyle{unsrt}

\bibliography{strings,refs}

\end{document}